\documentclass{article}

\PassOptionsToPackage{numbers, compress}{natbib}
\usepackage[final]{neurips_2022}
\usepackage{multirow}
\usepackage{bm}

\usepackage{multirow}
\usepackage{subfigure}
\usepackage[utf8]{inputenc} 
\usepackage[T1]{fontenc}    
\usepackage{hyperref}       
\usepackage{url}            
\usepackage{booktabs}       
\usepackage{amsfonts}       
\usepackage{nicefrac}       
\usepackage{microtype}      
\usepackage{xcolor}         
\usepackage{graphicx}
\usepackage{makecell}
\usepackage{wrapfig,lipsum,booktabs}
\usepackage{enumitem}

\title{A Large Scale Search Dataset for \\ Unbiased Learning to Rank}
\author{%
Lixin Zou$^1$, Haitao Mao$^2$\thanks{Work conducted during an internship at Baidu Inc.}\,\,\thanks{Equal contribution.}\,\,, Xiaokai Chu$^{1, \ast}$, Jiliang Tang$^2$, \\
\textbf{Shuaiqiang Wang}$^1$, \textbf{Wenwen Ye}$^1$, \textbf{Dawei Yin}$^1$\thanks{Corresponding author.} \\
$^1$Baidu Inc., $^2$Michigan State University \\
\texttt{\{zoulixin15,xiaokaichu, two\_ye, shqiang.wang\}@gmail.com} \\
\texttt{\{haitaoma,tangjili\}@msu.edu},
\texttt{yindawei@acm.org}
}

\begin{document}

\newcommand{\myauthornote}[3]{{\color{#2} {\sc #1}: #3}}
\newcommand{\authorText}[2]{{\color{#1}#2}}
\newcommand{\dataset}{Baidu-ULTR}

\newtheorem{defn}{Definition}

\maketitle
\begin{abstract}
The unbiased learning to rank (ULTR) problem has been greatly advanced by recent deep learning techniques and well-designed debias algorithms. However, promising results on the existing benchmark datasets may not be extended to the practical scenario due to some limitations of existing datasets. First, their semantic feature extractions are outdated while state-of-the-art large-scale pre-trained language models like BERT cannot be utilized due to the lack of original text. Second, display features are incomplete; thus in-depth study on ULTR is impossible such as the displayed abstract for analyzing the click necessary bias. Third, synthetic user feedback has been adopted by most existing datasets and real-world user feedback is greatly missing. To overcome these disadvantages, we introduce the Baidu-ULTR dataset. It involves randomly sampled 1.2 billion searching sessions and 7,008 expert annotated queries~(397,572 query document pairs). Baidu-ULTR is the first billion-level dataset for ULTR. Particularly, it offers: (1)~the original semantic features and pre-trained language models of different sizes; (2)~sufficient display information such as position, displayed height, and displayed abstract, enabling the comprehensive study of multiple displayed biases; and (3)~rich user feedback on search result pages (SERPs) like dwelling time, allowing for user engagement optimization and promoting the exploration of multi-task learning in ULTR. Furthermore, we present the design principle of Baidu-ULTR and the performance of representative ULTR algorithms on Baidu-ULTR. The Baidu-ULTR dataset and corresponding baseline implementations are available at \url{https://github.com/ChuXiaokai/baidu_ultr_dataset}. The dataset homepage is available at \url{https://github.com/HaitaoMao/baidu_ultr_page}.

\end{abstract}
\section{Introduction}
Learning to Rank~(LTR) that aims to measure documents' relevance w.r.t. queries is a popular research topic with applications in web search engines, e-commerce, and multiple different streaming services~\cite{qin2010letor}. 
With the rise of deep learning, the heavy burden of data annotation drives the academia and industry communities to the study of learning to rank using implicit user feedback~(e.g., user click).
However, directly optimizing the model with click data results in unsatisfied performance due to the existence of biases, such as position bias~\cite{joachims2017unbiased}, trust bias~\cite{agarwal2019addressing}, and click necessary bias~\cite{mao2018constructing}. 
Unbiased learning to rank~(ULTR) has been proposed for mitigating biases in user feedback with counterfactual learning algorithms~\cite{joachims2017unbiased, wang2016learning,wang2018position}. 
To meet the demand of ULTR, numerous datasets have been released publicly such as ~Yahoo! LETOR~\cite{chapelle2011yahoo}, Microsoft LETOR~\cite{qin2010letor}, Istella LETOR~\footnote{http://quickrank.isti.cnr.it/istella-dataset/}~\cite{dato2016fast}, and Tiangong-ULTR~\footnote{http://www.thuir.cn/data-tiangong-ultr/}~\cite{Ai2018ULR, Ai2018unbiased}. 
These datasets include thousands of annotated queries, hundreds of semantic features, and millions of user sessions~(from Tiangong), which have been extensively adopted for the ULTR research~\cite{ai2021unbiased}. 

Although existing ULTR datasets can meet the data-consuming requirement to some extent, there are still some limitations. 
First, the provided semantic features~(e.g., BM25~\cite{robertson1994some, robertson1995okapi, robertson2009probabilistic}, TF-IDF~\cite{salton1988term}, LMABS~\cite{zhai2017study}) cannot enjoy the advantages of modern representation learning techniques. 
Applying the recent paradigm of large-scale pre-training and end-to-end finetuning~(e.g., RoBERTa~\cite{liu2019roberta}, ERNIE~\cite{zhang2019ernie}, Poly-Encoder\cite{humeau2019poly}) requires the access to the original text of queries and documents. 
Second, limited types of display information are provided. 
The ranking position from the Tiangong-ULTR dataset is the only publicly available display information.
As expected, position-related biases~(e.g., position bias~\cite{joachims2017unbiased}, trust bias~\cite{agarwal2019addressing}) have been widely studied. However, biases related to other display information are overlooked. 
For example, users unnecessarily click the document if the document displayed abstract can perfectly meet user requirement~\cite{mao2018constructing}. 
Moreover, documents with different multimedia types could have a distinct attraction to users, e.g., videos and pictures are more attractive than plain text.
Third, most existing datasets lack real-world user feedback. Though Tiangong-ULTR provides real-world click data, a small test set with only 100 queries makes it difficult to produce reliable and significant results.
Alternatively, the research community seeks to simulate click data with the clicking behavior assumption that is consistent with the proposed methods~\cite{Ai2018ULR,jin2020deep,Jrvelin2017IREM}.
Though significant improvement has been observed in simulations, such success is hard to be extended to the practical scenario~\cite{ai2021unbiased,Ai2018ULR,DBLP:conf/www/ChapelleZ09}.

To address the aforementioned challenges, we introduce the \dataset{} dataset, a large-scale unbiased learning to rank dataset for web search. 
It randomly sampled 1.2 billion searching sessions from the largest Chinese search engine -- Baidu, and 7,008 expert annotated queries~(397,572 query document pairs) for validation and test. Overall, \dataset{} has the following advantages over existing ULTR datasets:
\begin{itemize}
    \item \dataset{} provides the original text of queries and documents after desensibilisation. 
    It enables us to construct both handcraft features and semantic features generated by advanced language models. In addition, we provide advanced language models pretrained with the MLM loss~\cite{devlin2018bert}.
    \item \dataset{} offers diverse display information such as position, displayed height, and so on.
    It enables the study of multiple biases with advanced techniques like causal discovery.
    \item Rich user behaviors, e.g., click, skip, dwelling time, and displayed time, have been recorded, offering opportunities for optimizing user engagement and exploring multi-task learning in ULTR.
    \item \dataset{} is a large-scale web search dataset~(1.2 billion searching sessions) with sufficient expert annotations~(397,572 expert-annotated query document pairs), which further supports pre-training large-scale language models, and  studies on the pre-training task for web search.
    \item \dataset{} introduces a more practical scenario with significant challenges in ULTR like ranking with long-tail queries and 
    conquering the mismatch between training and test sets.
     
\end{itemize}

The rest of the paper is organized as follows. In Section~\ref{sec:preliminary}, we briefly review the ULTR task and existing datasets. In Section~\ref{sec:dataset_description}, we give a detailed introduction on the dataset collection and data analysis. In Section~\ref{sec:experiments}, we conduct empirical studies with state-of-the-art ULTR methods. Finally, we discuss the impact and potential limitations of \dataset{} in Section~\ref{sec:discussion}.

\section{Preliminary}\label{sec:preliminary}
In this section, we briefly introduce the ULTR task and review the existing ULTR datasets with a detailed comparison with \dataset{} in Tab.~\ref{tab:dataset_summary}.

\subsection{Unbiased Learning to Rank}
The task of ranking is to measure the relative order among a set of $N$ documents $\mathcal{D}_q = \left\{d_i\right\}_{i=1}^N$ under the constraint of a query $q\in \mathcal{Q}$, where $\mathcal{D}_q \subset \mathcal{D}$ is the set of $q$-related documents retrieved from all indexed documents $\mathcal{D}$, and $\mathcal{Q}$ is the set of all possible queries. 
We aim to design a scoring function $f(q,d): \mathcal{Q}\times\mathcal{D} \rightarrow \mathbb{R}$ that descendingly sorts the documents as a list $\pi_{f,q}$ to maximize an evaluation metric $\vartheta$~(e.g., DCG~\citep{Jrvelin2017IREM}, PNR~\cite{zou2021pre}, and ERR~\cite{chapelle2011yahoo}) as
\begin{eqnarray}
f^\ast = \max_{f} \mathbb{E}_{q\in \mathcal{Q}}\vartheta(\mathcal{R}_{q},\pi_{f,q}).
\end{eqnarray}
where $\mathcal{R}_{q} = \{r_{d}\}_{d\in\mathcal{D}_{q}}$ is a set of relevance labels $r_d$ corresponding to $q, d$. Usually, $r_{d}$ is the graded relevance in 0-4 ratings, which indicates the relevance of document $d_i$ as $\{$\textbf{bad}, \textbf{fair}, \textbf{good}, \textbf{excellent}, \textbf{perfect}$\}$, respectively.
To learn the scoring function $f$, a corresponding loss function is needed to approximate $r_d$ with $f(q,d)$ as
\begin{equation}
    \label{eq:ideal_loss}
    \ell_{ideal}(f) =  \mathbb{E}_{q\in \mathcal{Q}}\left[  \sum_{d\in\mathcal{D}_q}\Delta(f(q,d),r_d)\right],
\end{equation}
where $\Delta$ is a function that computes the individual loss for each document. 
If all the documents are annotated, $\ell_{ideal}(f)$ would be the ideal ranking loss for optimizing the ranking function. 
Typically, the relevance annotation $r_{d}$ is elicited through expert judgment; thus $r_d$ is considered to be unbiased, but expensive. 

An alternative but intuitive approach is to use the user's implicit feedback as the relevance label. For example, by replacing the relevance label $r_d$ with click label $c_d$ in Equ.~\ref{eq:ideal_loss}, a naive empirical ranking loss is derived as follows:
\begin{eqnarray}
    \label{eq:loss_naive}
    \ell_{naive}(f) = \frac{1}{|\mathcal{Q}_{o}|}\sum_{q\in \mathcal{Q}_{o}}\left[  \sum_{d\in\mathcal{D}_q}\Delta\left(f(q,d),c_d\right)\right],
\end{eqnarray}
where $\mathcal{Q}_{o}$ is the \textit{observed} query set. 
$c_{d}$ is a binary variable indicating whether the document $d$ in the ranked list is clicked or not. 
However, this naive loss function is biased since the display of the final ranking result may influence user's click~\cite{wang2016beyond}. 
For instance, position bias occurs because users are more likely to examine the documents at higher ranks~\cite{joachims2017unbiased}. 
Consequently, highly ranked documents may receive more clicks, and relevant (but unclicked) documents may be perceived as negative samples because they are unexamined by users. 
To address this issue, unbiased learning to rank has been proposed to remove the effect of data bias in the computation of the ranking loss $\ell_{naive}(f)$.
so that the model trained with biased data (e.g., clicks) would converge to that trained with unbiased labels (i.e., the relevance of a document).

\subsection{Existing ULTR Datasets} 
Existing publicly available ULTR datasets can be roughly categorized by utilizing synthetic or real user feedback. Both have been widely adopted by the empirical study of ULTR algorithms.

\paragraph{Synthetic Data}
Yahoo! LETOR~\cite{chapelle2011yahoo}, Microsoft LETOR~\cite{qin2010letor} and Istella LETOR~\cite{dato2016fast} are three commonly used datasets with synthetic user feedback. 
Notice that Yahoo! LETOR includes two sets, Yahoo Set1, and Yahoo Set2.
Due to privacy concerns, those datasets do not release real user feedback and hide the original text of queries and documents. Thus, researchers have to simulate click data following a specific user behavior assumption, such as position-dependent click model~\cite{joachims2017unbiased}, and cascade click model~\cite{craswell2008experimental}. 
Furthermore, a set of predefined semantic features is used to represent the original queries and documents. Detailed statistics about those datasets are illustrated in Tab.~\ref{tab:dataset_summary}.

\paragraph{Real Data} 
The Tiangong-ULTR~\cite{Ai2018ULR, Ai2018unbiased} is the only dataset with real-world user feedback  supporting the research of unbiased learning to rank. 
It provides real-world click data sampled from the search sessions of Sogou.com~\footnote{https://www.sogou.com/} for training and a separate expert annotated test set for the performance evaluation.
However, the test set with only 100 queries is insufficient to draw a significant conclusion with the dataset. 
Detailed statistics about Tiangong-ULTR can be found in Tab.~\ref{tab:dataset_summary}. 

\begin{table}[h]
\small
\tabcolsep 0.02in
\renewcommand{\arraystretch}{1.2}
\centering
\caption{Characteristics of publicly available datasets for unbiased learning to rank.}
\scalebox{0.77}{
\begin{tabular}{l r r l l r r r r r c}
\bottomrule
& \multicolumn{5}{c}{Training Implicit Feedback Data} & \multicolumn{3}{c}{Validation \& Test Data} &  \\ \cmidrule(lr){2-6}  \cmidrule(lr){7-9}
Dataset & \# Query & \# Doc & \# User Feedback & \# Display-info & \# Session & \# Query &  \# Doc  &\# Label & \# Feature & Pub-Year  \\\hline
Yahoo Set1 & 19,944 & 473,134 & 1~(Simulated click) & 1~(Position) & - & 9,976 & 236,743 & 5 & 519 & 2010 \\
Yahoo Set2 & 1,266 & 34,815 & 1~(Simulated click) & 1~(Position) & - & 5,064 & 138,005 & 5 & 596 & 2010 \\
Microsoft & $\approx$18,900 & $\approx$2,261,000 & 1~(Simulated click) & 1~(Position) & - & $\approx$12,600 & $\approx$1,509,000 & 5 & 136 & 2010 \\
Istella & 23,219 & 7,325,625 & 1~(Simulated click) & 1~(Position) & - & 1,559 & 550,337 & 5 & 220 & 2016 \\
Tiangong & 3,449 & 333,813 & 1~(Real Click) & 1~(Position) & 3,268,177 & 100 & 10,000 & 5 & 33 & 2018 \\
Baidu & 383,429,526 & 1,287,710,306 & 18~(Real Feedback) & 8~(Display Info) & 1,210,257,130 & 7,008 & 367,262 & 5 & ori-text & 2022 \\
\toprule
\end{tabular}
}
\label{tab:dataset_summary}
\end{table}

\section{Dataset Description}\label{sec:dataset_description}
In this section, we formally introduce the \dataset{} dataset. It consists of two parts: \textbf{(1)} Large Scale Web Search Sessions and \textbf{(2)} Expert Annotation Dataset. Next, we will first detail these two parts. Then, we provide detailed data analysis in Section~\ref{sec:data_analysis} for a better understanding of the collected dataset.
The data license is further provided in Section~\ref{app:license}.

\subsection{Large Scale Web Search Sessions \label{sec:trainset}}
\paragraph{Queries}\label{sec:query_collection} Queries are randomly sampled from search sessions of the Baidu search engine in April 2022.
Each query is given a unique identifier.
A frequent query is likely to be sampled several times from search sessions, and each replicate is given a different identifier. 
This ensures the query distribution in \dataset{} follows the same distribution as that in the online system: frequent queries have larger weights.

\paragraph{Documents} 
 For the candidate documents of the query, we only record the displayed documents to save the cost of storing the logs. Typically, the document not being displayed is less informative than the  ones displayed since the user cannot provide any feedback on the document without display.
 As a result, the logged search session usually contains just 10 documents for every query because one page contains 10 results. Only 1.1\% search sessions contain over 10 displayed documents, i.e., only 1.1\% of users turn the page. This phenomenon further results in the mismatch between the training and test.   
 Specifically, in the training phase, users just ‘‘label'' the top-10 results. However, in the inference stage, the top-10 results are generated through \textbf{retrieval} and \textbf{ranking}, where the systems are required to rank billion documents or thousands of documents by multiple models such as Inverted index~\cite{yan2009inverted}, BM25~\cite{robertson2009probabilistic}, TF-IDF~\cite{aizawa2003information}, Bi-Encoder~\cite{humeau2019poly}, RankNet~\cite{burges2010ranknet}, BM25, DSSM~\cite{wan2016deep} and Cross-encoder~\cite{zou2021pre}).

\begin{figure}
    \centering
      \scalebox{0.34}{\includegraphics{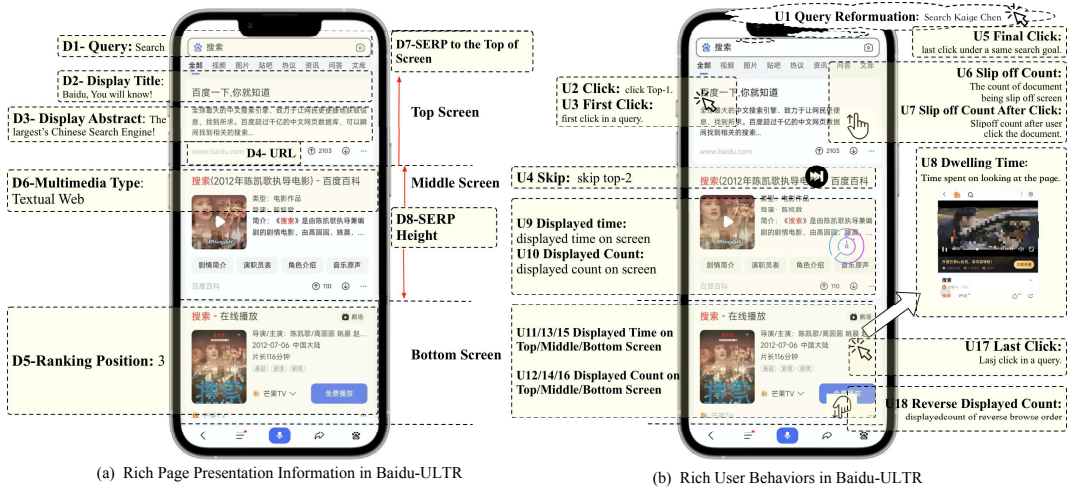}}
    \caption{
    (a) A demo explanation of rich page presentation information in \dataset{}. There are 8 presentation features that start from D1 to D8. Detailed descriptions are illustrated in Tab.~\ref{tab:page_presentation} in Appendix~\ref{app:data_description}. 
    (b) A demo explanation of rich user behaviors in \dataset{}. There are 18 user behaviors starting from U1 to U18. Detailed descriptions are illustrated in Tab.~\ref{tab:user_behavior} in Appendix~\ref{app:data_description}. 
    }
    \label{fig:display_behavior_dataset}
\end{figure}

\paragraph{Page Presentation Features} 
User behaviors on search result pages (SERPs) are highly biased by the layout of SERPs~\cite{wang2016beyond}, such as the position~(position bias~\cite{joachims2017unbiased}), the displayed abstract of the document~(click necessary bias~\cite{mao2018constructing}), and the displayed area of the document. 
To support advanced research in ULTR, we collect the following rich presentation information of the document on SERPs: the ranking \textbf{position}, the displayed \textbf{url}, the displayed \textbf{title} of the document, the displayed \textbf{abstract} of the document, the \textbf{multimedia type} of the document, and the height of SERP~(i.e., the vertical pixels of SERP on the screen). 
An illustration of the page presentation features is shown in Fig.~\ref{fig:display_behavior_dataset}(a). 
The type and description of presentation features are illustrated in Tab.~\ref{tab:page_presentation} in Appendix~\ref{app:data_description}. 

For user privacy protection, \textbf{the original texts of query, title and abstract are denoted as sequential token ids with a private dictionary}. 
To further prevent losing semantic information, we provide a unigram set that records the high-frequency words using the desensitization token ids, which is useful for modeling the word-level semantic information. 
For easy usage, we provide a set of pre-trained language models with various sizes~(pre-trained with MLM and naive loss mentioned in Equ.~\ref{eq:loss_naive}) that can translate the raw text into dense semantic features.

\paragraph{User Behaviors} 
Due to the lack of user behaviors, ULTR research mainly focuses on click modeling~\cite{ai2021unbiased} and few industrial works analyze the user's satisfaction with rich user behaviors~\cite{wang2016beyond}.
To facilitate the research on ULTR, we provide a set of rich users' behaviors, including user's \textbf{query reformulation}, the \textbf{skip}, the \textbf{click}, the \textbf{first click}, the \textbf{last click}, user's \textbf{dwelling time}, the \textbf{displayed time} on the screen, the \textbf{displayed count} on the screen,  the \textbf{slip off count}, and the displayed count of reverse browsing. 
A demo explanation is presented in Fig.~\ref{fig:display_behavior_dataset}(b).
The type and description of user behaviors are illustrated in Tab.~\ref{tab:user_behavior} in Appendix~\ref{app:data_description}.

\subsection{Expert Annotation Dataset}

\begin{figure}
    \centering
    \includegraphics[width=1.01\textwidth]{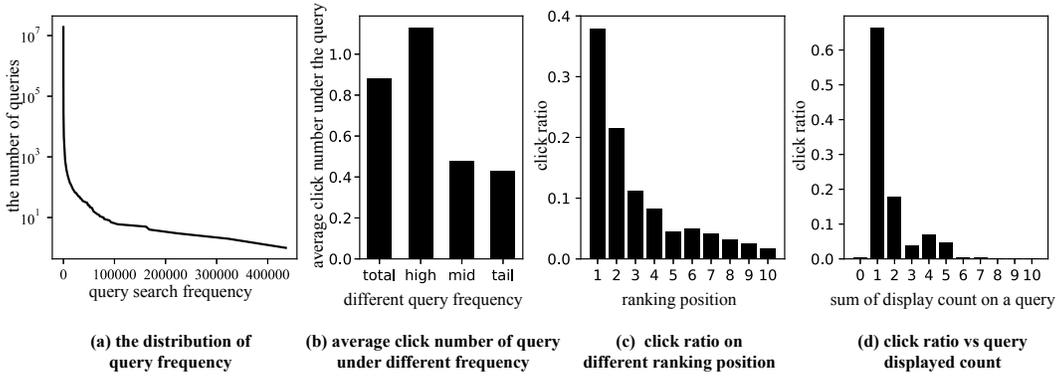} 
    \caption{Data analysis on Baidu-ULTR: (a) the distribution of query frequency. (b) the average click number of queries under different frequencies. (c) the click ratio on different ranking positions. (d) the click ratio vs the query displayed count (the sum of display count on a query).}\label{fig_data_analysis}
\end{figure}

\paragraph{Queries} 
Queries are randomly sampled from the monthly collected query sessions of the Baidu search engine, which is similar to the query collection in Section~\ref{sec:query_collection}. 
Since the search queries are heavy-tailed distributed~(depicted in Fig.~\ref{fig_data_analysis}(a)), we further provide the frequency of queries. 
Specifically, queries are descendingly split into 10 buckets according to their monthly search frequencies, 
where buckets 0, 1, 2, buckets 3, 4, 5, 6, and buckets 7, 8, 9 correspond to the high-frequency, mid-frequency, tail frequency, respectively. 
The rankers' performance w.r.t. the queries frequency can be further analyzed, which is beneficial for
understanding the model's strengths and weaknesses under the long-tail phenomenon.

\paragraph{Documents} To simulate the online scenario, candidate documents are selected from the retrieval phase. 
Specifically, we record the top-30 documents from the retrieval model as the candidate set for ranking. We further cover the top thousands of results from the retrieval stage by selecting the document with an interval of 30~(i.e., the documents ranked at position $\{30, 60, 90, 120, 150, ..., 990\}$). These documents are used to measure rankers' performance on distinguishing documents that are extremely disruptive to the user experience.

\begin{wraptable}{r}{8cm}
\tabcolsep 0.05in
    \centering
    \caption{Distribution of Relevance Labels.}
    \scalebox{0.9}{
    \begin{tabular}{lc|rr}
    \bottomrule
      Grade & Label &\# Query-Doc & Ratio of Label \\\hline
      Perfect & 4 & 714 & 1.80\% \\
      Excellent & 3 & 28,172 & 9.21\%\\
      Good & 2 & 112,759 & 28.36\%\\
      Fair & 1 & 36,622 & 9.21\%\\
      Bad & 0 & 219,305 & 55.16\% \\\toprule
    \end{tabular}}
    \label{tab:relevance_distribution}
\end{wraptable} 

\paragraph{Expert Annotation} The relevance of each document to the query has been judged by expert annotators who assign one of 5 labels, $\{$\textbf{bad}, \textbf{fair}, \textbf{good}, \textbf{excellent}, \textbf{perfect}$\}$ to the document. 
Each of these relevance labels is then converted to an integer ranging from 0 (for bad) to 4 (for perfect). 
Specific guidelines, depicted in Tab.~\ref{tab:guideline}, are given to expert annotators for instructing them on how to perform relevant judgments.
The main purpose of these guidelines is to reduce the amount of disagreement across expert annotators. Additionally, only the label that most expert annotators agree on will be selected as the final label.
Tab.~\ref{tab:relevance_distribution} illustrates the distribution of the relevance labels. We can observe: \textbf{(1)} perfect only occupies 1.8\% in all expert annotations since a perfect will be only given to the destination page of a navigational query according to Tab.~\ref{tab:guideline}; \textbf{(2)} the bad documents take over 50\% of documents. The reason is that the irrelevant documents are usually the majority for long-tail queries, e.g., ``Does the China Tobacco Bureau have any temporary job position?'' Unfortunately, the long-tail queries are also the majority for the search engine~(referred to Fig.~\ref{fig_data_analysis}(a)).

\begin{table}[h]
\tabcolsep 0.02in
\centering
\caption{The general guideline of annotation.}
\begin{tabular}{l|l}
\bottomrule
Label & Guideline  \\\hline
0 (bad) & Useless or outdated documents that do not meet the requirements at all.\\
1 (fair) & Helpful to some extent but deficient in authority, timeliness document. \\
2 (good) & Meet the requirement of the query.\\
3 (excellent) & Meet the requirement of the query and timeliness document. \\
4 (Perfect) & Meet the requirement of the query, timeliness, and authoritative document.  \\
\toprule
\end{tabular}
\label{tab:guideline}
\end{table}

\subsection{Dataset Analysis \label{sec:data_analysis}} 
In this subsection, we present our primary data analysis on the \dataset{} dataset. A full analysis procedure and more detailed results can be found in Appendix~B. As shown in Fig.~\ref{fig_data_analysis}, we have the following observations:

\begin{enumerate}
    \item Long-tail distribution appears in many user behaviors.
    As shown in Fig.~\ref{fig_data_analysis}(a), over 60\% searches are based on top 10\% high-frequency queries, while most queries only appear very few times. This phenomenon also happens on the displayed count, the displayed time, the number of clicks per query, and the number of skips per query~(depicted in Appendix~B).
    The long tail distribution can significantly affect the ULTR performance. As reported in Tab. \ref{tab:search_frequency}, all ULTR algorithms perform poorly on those tail queries.
    \item The logged search results on the high-frequency queries are more relevant than those on tail queries since the number of clicks per query on the high-frequency queries with an average click of 1.13 is much larger than those on tail queries with an average click of 0.43 (shown in Fig.~\ref{fig_data_analysis}(b)).
    \item The most critical user feedback (i.e., click) shows a strong correlation with both page presentation features, e.g., position, and other user behaviors, e.g, displayed count. 
    We illustrate the click ratio on different ranking positions in Fig.~\ref{fig_data_analysis}(c). 
    The click ratio decreases as the documents are displayed in the latter positions. 
    Similar observations can also be made in click and displayed count. As shown in Fig.~\ref{fig_data_analysis}(d),
    more than 70\% of clicks are done with only one displayed count on the document belonging to the query.
    \item As the displayed time becomes longer, users are more likely to spend more time on the top of the screen, from 25\% to nearly 35\%, while less time is spent on the bottom of the screen from more than 40\% to 30\% (depicted in Fig.~\ref{fig:app_different_screen}).
\end{enumerate}

\subsection{\dataset{} License}\label{app:license}
The dataset can be freely downloaded at~\url{https://github.com/ChuXiaokai/baidu_ultr_dataset} and noncommercially used with a custom license CC BY-NC 4.0\footnote{\url{https://creativecommons.org/licenses/by-nc/4.0/}}. Besides the current tasks in the dataset directory, users can define their own ones under the license.

\section{Benchmark and Baselines}\label{sec:experiments}
In this section, we conduct an empirical study of several benchmark unbiased learning to rank algorithms on \dataset{} and further present their performance versus different query frequencies.

\subsection{Baseline Methods}
To fully test unbiased learning to rank algorithms with different learning paradigms, we select the following representative baselines:

\begin{itemize}
  \item \textbf{Naive}: It directly trains the model with user feedback without any correction, as stated in Equ.~\ref{eq:loss_naive}.
  \item \textbf{IPW}: Inverse Propensity Weighting is one of the first ULTR algorithms proposed under the framework of counterfactual learning~\cite{joachims2017unbiased}, which weights the training loss with the probability of the document being examined in the search session.
  \item \textbf{DLA}: The Dual Learning Algorithm~\cite{Ai2018ULR} treats the problem of unbiased learning to rank and unbiased propensity estimation as a dual problem, such that they can be optimized simultaneously.  
  \item \textbf{REM}: The Regression EM model~\cite{wang2018position} uses an EM framework to estimate the propensity scores and ranking scores. 
  \item \textbf{PairD}: The Pairwise Debiasing Model~\cite{DBLP:conf/www/HuWPL19} uses inverse propensity weighting for pairwise learning to rank. 
\end{itemize}

Notably, all the above ULTR algorithms only take the position-related bias into consideration without utilizing other display features. Moreover, they only include the click data as the supervised signal without any other user behavior data. Therefore, there are great potentials for the design of the new ULTR algorithms on Baidu-ULTR dataset by utilizing the new extracted SERPs and display features.

\subsection{Metrics}
The following evaluation metrics are employed to assess the performance of the ranking system.
The \textbf{Discounted Cumulative Gain} (DCG)~\citep{Jrvelin2017IREM} is a standard listwise accuracy metric and is widely adopted in the context of ad-hoc retrieval. 
For a ranked list of $N$ documents, we use the following implementation of DCG:
\begin{eqnarray*}
    DCG@N = \sum_{i=1}^N \frac{G_i}{\log_2(i+1)},
\end{eqnarray*}
where $G_i$ represents the weight assigned to the document's label at position $i$.
A higher degree of relevance corresponds to a higher weight.
We use the symbol $DCG$ to indicate the average value of this metric over the test queries. 
$DCG$ will be reported only when absolute relevance judgments are available. 

The \textbf{Expected Reciprocal Rank}~(ERR)~\cite{ghanbari2019err} calculates the expectation of the reciprocal of the position of a result at which a user stops. 
This measure is defined as:
\begin{eqnarray*}
    ERR@{N}=\sum_{i=1}^{N} \frac{1}{i} \prod_{j=1}^{i-1}\left(1-R_j\right) R_i,
\end{eqnarray*}
where $R_i$ indicates the relevance probability of the $i$-th document to the query and the expression $\frac{1}{i} \prod_{j=1}^{i-1}\left(1-R_j\right)$ represents the non-relevance probability of the ordered documents prior to the position of the $i$-th document in the list.

\subsection{Model Setup}
In our experiments, we employ a transformer-based cross-encoder with 12 layers, 12 heads, and 768 hidden size as the backbone model for approximating the scoring function $f(q,d)$. 
For easy usage, we provide a warm-up model for initialization, which is trained with the mixture of the naive loss~(Equ.~\ref{eq:loss_naive}) and the masked language modeling~(MLM) loss~\cite{devlin2018bert} by randomly masking 10\% tokens. With the pre-trained model, the ranking function is set by using the \texttt{[CLS]} embedding with 3-layers MLP~(hidden layer of 512-256-128) and optimized with the Adam optimizer~(learning rate = $2\times 10^{-6}$). 
All the experiments are obtained by an average of 5 repeat runs.
The hyper-parameters for training the ULTR algorithms are selected from ULTR toolkit\footnote{\url{https://github.com/ULTR-Community/ULTRA_pytorch}}~\cite{tran2021ultra}. 
The expert annotation dataset is split into validation and test sets according to a 20\%-80\% scheme.
All the models are trained on the machine with $28$ Intel(R) 5117 CPU, $32G$ Memory, 8 NVIDIA V100 GPUs, and 12T Disk.

\subsection{Performance Comparison}
Tab.~\ref{tab:experimental_result} summarizes the DCG and ERR performance of selected learning algorithms on the \dataset{} dataset. From the table, we have the following observations: 
(1) No ULTR algorithms provide satisfying results since they only take the position-related biases into consideration. 
(2) The DLA algorithm performs best across all algorithms, showing its robustness on the real dataset. (3) The naive algorithm shows comparable performance with IPW, even better than REM and PairD.
It reveals that real-world user feedback can be more complex than synthetic feedback generated with specific user behavior assumptions like position-dependent click model~\cite{joachims2017unbiased}. 
Therefore, ULTR algorithms with good performance on synthetic datasets may not show consistently good performance in the real-world scenario. An intuitive explanation is that the user logs are collected from a well-performed search engine, the click may not have such bias as mentioned in the assumption. Directly mitigating the bias may lead to over-mitigation and unsatisfying results.

\begin{table}[h]  
    \caption{Comparison of unbiased learning to rank (ULTR) algorithms with different learning paradigms on \dataset{} using cross-encoder as ranking models. The best performance is highlighted in boldface.}
    \centering
    \tabcolsep 0.03in
    \renewcommand{\arraystretch}{1.2}
    \scalebox{0.82}{
    \begin{tabular}{l|cccccccc}
    \bottomrule
    & DCG@1 & ERR@1 & DCG@3 & ERR@3 & DCG@5 & ERR@5 & DCG@10 & ERR@10\\\hline
    Naive & 1.235$\pm$0.029 & 0.077$\pm$0.002 & 2.743$\pm$0.072 & 0.133$\pm$0.003 & 3.889$\pm$0.087 & 0.156$\pm$0.003 & 6.170$\pm$0.124 & 0.178$\pm$0.003\\
    IPW & 1.239$\pm$0.038 & 0.077$\pm$0.002 & 2.742$\pm$0.076 & 0.133$\pm$0.003 & 3.896$\pm$0.100 & 0.156$\pm$0.004 & 6.194$\pm$0.115 & 0.178$\pm$0.003\\
    REM & 1.230$\pm$0.042 & 0.077$\pm$0.003 & 2.740$\pm$0.079 & 0.132$\pm$0.003 & 3.891$\pm$0.099 & 0.156$\pm$0.004 & 6.177$\pm$0.126 & 0.178$\pm$0.004\\
    PairD & 1.243$\pm$0.037 & 0.078$\pm$0.002 & 2.760$\pm$0.078 & 0.133$\pm$0.003 & 3.910$\pm$0.092 & 0.156$\pm$0.003 & 6.214$\pm$0.114 & 0.179$\pm$0.003 \\
    DLA & \textbf{1.293}$\pm$0.015 & \textbf{0.081}$\pm$0.001 & \textbf{2.839}$\pm$0.011 & \textbf{0.137}$\pm$0.001 & \textbf{3.976}$\pm$0.007 & \textbf{0.160}$\pm$0.001 & \textbf{6.236}$\pm$0.017 & \textbf{0.181}$\pm$0.001 \\
    \toprule
    \end{tabular}}
    \label{tab:experimental_result}
\end{table}

\subsection{Performance Comparison on Tail Query}
To investigate more on the effect of search frequencies, we further compare the ULTR algorithm performance versus three different levels of search frequencies, including high, middle, and tail.
Results in Tab.~\ref{tab:search_frequency} demonstrate the following observations: (1) performance of all the algorithms decreases progressively from high frequency queries to tail queries, which indicates the difficulty of learning to rank on the tail queries; (2) The naive algorithms show competitive performance on the tail queries, revealing that the benchmark methods are fragile in dealing with the data bias in tail queries. 
(3) The DLA algorithm illustrates considerable improvement in the high-frequency queries, suggesting that the ULTR algorithms are beneficial for the high-frequency ranking problem.

\begin{table}[h]
\centering
\small
\renewcommand{\arraystretch}{1.08}
\tabcolsep 0.03in
\caption{Performance comparison of evaluation ULTR algorithms versus different search frequencies. The best performance is highlighted in boldface.}
\renewcommand{\arraystretch}{1.2}
\scalebox{0.8}{
\begin{tabular}{l|ccc|ccc|ccc}
\bottomrule
\multirow{2}{*}{Model} & \multicolumn{3}{c|}{DCG@3} & \multicolumn{3}{c|}{DCG@5} & \multicolumn{3}{c}{DCG@10} \\
& High & Mid & Tail & High & Mid & Tail & High & Mid & Tail \\ \hline
Naive & 3.960$\pm$0.058 & 2.992$\pm$0.119 & 1.742$\pm$0.079 & 5.596$\pm$0.098 & 4.254$\pm$0.142 & \textbf{2.474}$\pm$0.092 & 8.812$\pm$0.140 & \textbf{6.777}$\pm$0.173 & 3.942$\pm$0.121 \\
IPW & 4.017$\pm$0.132 & 2.976$\pm$0.111 & 1.722$\pm$0.061 & 5.699$\pm$0.145 & 4.235$\pm$0.140 & 2.447$\pm$0.090 & 8.969$\pm$0.146 & 6.762$\pm$0.163 & 3.925$\pm$0.109 \\
REM & 3.994$\pm$0.114 & 2.982$\pm$0.124 & 1.723$\pm$0.067 & 5.665$\pm$0.128 & 4.237$\pm$0.158 & 2.454$\pm$0.074 & 8.904$\pm$0.147 & 6.755$\pm$0.183 & 3.927$\pm$0.104 \\
PairD & 4.018$\pm$0.102 & 2.993$\pm$0.110 & \textbf{1.750}$\pm$0.079 & 5.662$\pm$0.120 & 4.267$\pm$0.129 & 2.474$\pm$0.088 & 8.924$\pm$0.145 & 6.804$\pm$0.153 & \textbf{3.961}$\pm$0.119 \\
DLA & \textbf{4.226}$\pm$0.042 & \textbf{3.073}$\pm$0.022 & \textbf{1.750}$\pm$0.016 & \textbf{5.894}$\pm$0.030 & \textbf{4.300}$\pm$0.020 & 2.472$\pm$0.009 & \textbf{9.147}$\pm$0.044 & 6.767$\pm$0.027 & 3.920$\pm$0.009 \\
\toprule
\end{tabular}
}
\label{tab:search_frequency}
\end{table}
\section{Discussion}\label{sec:discussion}
In this section, we will discuss 
(1) the challenges of designing well-performed algorithms on our \dataset{};
(2) the new research topics introduced to the academic study;
(3) the potential limitations existing in \dataset{}.

\subsection{Data Challenges}
In this subsection, we introduce three new challenges on how to debias in \dataset{}, which is more practical with a close connection to the industry scenarios.

\paragraph{Biases in Real-World User Feedback}
Different from other datasets, the user feedback and display information are collected from the user logs in Baidu search engine.
As the user logs are collected from a well-performed search engine, the click may not have the same simple and strong bias as the assumption in generating synthetic datasets, i.e. \cite{joachims2017unbiased,Ai2018ULR,Jrvelin2017IREM}.
Directly mitigating the bias may lead to over-mitigation and unsatisfying results.

\paragraph{Long-tail Phenomenon}
Long-tail phenomenons happen frequently in the \dataset{} as shown in Fig. \ref{fig_data_analysis}. 
We selectively emphasize two important long-tail phenomenons as follows. 
(1) Most retrieved documents are irrelevant to the user query as shown in the expert annotation. 
Over 50\% of documents are annotated as irrelevant with the user query while only 1.8\% of documents are annotated as perfect.
(2) The user query also shows the long-tail distribution. The top 10\% high-frequency queries occupy over 60\% of search logs.
In Tab. \ref{tab:search_frequency}, the performance of ULTR algorithms on the high-frequency queries is much higher than the tail queries.

\paragraph{Mismatch between Training and Test} 
Due to the storage limitation in the online system, 
we only record the displayed pages in the ranking system, which is usually the top-10 results. 
However, in the evaluation set, we record the top-30 documents and further cover the top thousands of results from the retrieval stage by selecting the document with an interval of 30. The sample results indexes are \{30, 60, 90, 120, 150, ..., 990\}.
This challenge widely exists in the practical scenario, an arbitrary number of documents can be retrieved into our ranking procedure. 
The number of documents for training and testing can seldom be exactly aligned. The data mismatching is an out-of-distribution challenge for ranking algorithms.

\subsection{Research Topics}
In this subsection, we introduce three advanced research topics with great practical value. 
\dataset{} first enables the academic studies of those topics in ULTR.  

\paragraph{Pre-training models for Ranking.} 
The pre-training learning paradigm has been proven to be beneficial for improving downstream task performance. 
However, pre-training for ranking has not been well explored yet. 
In \dataset{}, a large corpus of queries and documents has been provided, which opens the opportunity for exploring the pre-training task for LTR. 
Moreover, finetuning on the downstream task also plays an important role in the learning paradigm. 
How to fine-tune the pre-training model with biased user feedback can also be a promising research direction.   

\paragraph{Causal Discovery.} 
Due to missing real implicit feedback in the existing datasets, existing methods follow the paradigm of making assumptions about the clicking model and verifying the assumption in a simulation experiment, which might not be realistic. With \dataset{}, we are capable of discovering the clicking model from the dataset and extracting the relevance model from the casual discovery model \cite{eberhardt2017introduction}.  

\paragraph{Multi-task Learning.} 
Different tasks can benefit each other by giving complementary information or counteracting task-independent information, such as CTR and CVR prediction~\cite{ma2018entire} in the recommendation. 
With sufficient user behaviors, \dataset{} is suitable for studying the benefit provided by the multi-complementary tasks, such as dwelling time prediction, which might promote this direction. 

\subsection{Data Limitations}  
In this subsection, we introduce three potential limitations of \dataset{} that should be noticed when utilizing \dataset{}.

\paragraph{Inconvenient Model Sharing.} 
Due to the privacy issue, the original text is translated into token ids with a private dictionary. 
Furthermore, to avoid losing the semantic features, we provide a set of high-frequency words with the combination of tokens in \dataset{}. 
Though the desensitized original text is beneficial for extracting all kinds of semantic features for LTR easily, commonly used pre-trained language models trained with large corpus such as BERT and ERNIE cannot be used directly. 
We have to retrain models solely with the \dataset{} corpus. 
For easy usage, we provide pre-trained language models as described in Section~\ref{sec:experiments}.

\paragraph{User Diversity.}
\dataset{} is collected from the query sessions in the Baidu search engine, which majorly targets on the Chinese community.
This may induce the lack of user diversity and language diversity.
Statistically speaking, there are about 97\% queries in Chinese and about 2\%  queries in English.

\paragraph{Biases in Expert Annotation} 
Though we have designed a detailed 80-page guideline for annotation, designing the expert annotation guideline is a non-trivial task. There are still two potential biases in the expert annotation dataset: 
(1) To some extent, improving on the debiasing algorithms might not reflect on the test dataset since the 5-level annotation is typically more coarse than user feelings on the results. 
(2) In some special queries, the guideline might conflict with the relevance judgment from users' feedback since there is no guideline suitable for all queries. For example, in medical queries, the authority may be more important since the users are more concerned about correctness of the documents. However, in the general guidelines, the timeless document is judged as more relevant than the authoritative document. Though, we have an 80-page guideline that considers all the known exceptions, we cannot guarantee to cover all the special cases.

\section{Conclusion \label{sec:conclusion}}
This paper introduces a large-scale web search dataset, \dataset{} dataset.
It includes large-scale web search sessions for training with real-world user feedback, and 7,008 expert annotation queries~(397,572 query document pairs) for evaluation. 
The \dataset{} contains adequate display information and rich types of user feedback. 
Moreover, the raw textual feature is also provided after desensitization, which enables the utilization of more advanced language models.
A set of well-trained language models is also provided. 
We systematically introduce each component in the \dataset{} and the instruction on how to conduct the dataset. 
Empirical studies show the limitations of the existing ULTR algorithms and the great potential to develop new algorithms with \dataset{}. 
We also carefully consider the broader impact from various perspectives such as fairness, security, and harm to people. No apparent risk is related to our work.

\bibliographystyle{plain}
\bibliography{main}
\newpage
\section*{CheckList}
\begin{enumerate}
\item For all authors...
\begin{enumerate}
  \item Do the main claims made in the abstract and introduction accurately reflect the paper's contributions and scope? \answerYes{}
    
  \item Have you read the ethics review guidelines and ensured that your paper conforms to them? \answerYes{}
  
  \item Did you discuss any potential negative societal impacts of your work?  \answerYes{see in \ref{sec:conclusion}}
  
  \item Did you describe the limitations of your work? \answerYes{See Section~\ref{sec:discussion}}.

\end{enumerate}

\item If you are including theoretical results... 
\begin{enumerate}
    \item Did you state the full set of assumptions of all theoretical results?  \answerNA{}
    
    \item Did you include complete proofs of all theoretical results? \answerNA{}
    
\end{enumerate}

\item If you ran experiments...
\begin{enumerate}
    \item Did you include the code, data, and instructions needed to reproduce the main experimental results (either in the supplemental material or as a URL)? \answerYes{}
    
    \item Did you specify all the training details (e.g., data splits, hyperparameters, how they were chosen)? \answerYes{}
    
    \item Did you report error bars (e.g., with respect to the random seed after running experiments multiple times)? \answerYes{}
    
    \item Did you include the total amount of compute and the type of resources used (e.g., type of GPUs, internal cluster, or cloud provider)? \answerYes{}
\end{enumerate}

\item If you are using existing assets (e.g., code, data, models) or curating/releasing new assets...
\begin{enumerate}
    \item If your work uses existing assets, did you cite the creators? \answerYes{}
  
    \item Did you mention the license of the assets? \answerYes{See in Appendix A}
    
    \item Did you include any new assets either in the supplemental material or as a URL? \answerYes{}
  
    \item Did you discuss whether and how consent was obtained from people whose data you're using/curating? \answerYes{} 
  
    \item Did you discuss whether the data you are using/curating contains personally identifiable information or offensive content? \answerYes{}
\end{enumerate}

\item If you used crowdsourcing or conducted research with human subjects...
\begin{enumerate}
    \item Did you include the full text of instructions given to participants and screenshots, if applicable? \answerYes{}
  
    \item Did you describe any potential participant risks, with links to Institutional Review Board (IRB) approvals, if applicable? \answerYes{}
  
    \item Did you include the estimated hourly wage paid to participants and the total amount spent on participant compensation? \answerYes{}
\end{enumerate}
    
\end{enumerate}
\newpage
\appendix
\section{Detailed Dataset Description}

\subsection{Detailed Feature Description}\label{app:data_description}
In this subsection, we will provide a detailed description of the features of the large-scale web search session mentioned in Sec.  \ref{sec:trainset}. 
Types and explanations on page presentation features and user behaviors are shown in Fig. \ref{tab:page_presentation} and Fig. \ref{tab:user_behavior}, respectively. One thing needed to be clarified is that, the displayed time is not necessary to be the summation of the displayed time on top, middle and bottom. 
The reason is that if the document displayed height is too large, for example, half of the screen, the display time will be recorded by both the top and middle of the screen.
\textbf{\begin{table}[h]
    \caption{Rich Page Presentation Information in \dataset{}}
    \centering
    \tabcolsep 0.02in
\scalebox{0.85}{
    \begin{tabular}{l|l|l}
    \bottomrule
    Presentation Info & Feature Type  & Explanation \\\hline
    Query & Sequential desensitization tokens.  & The user issued query. \\\hline
    Displayed Title & Sequential desensitization tokens. & The title of document.  \\\hline
    Displayed Abstract & Sequential desensitization tokens. & A query-related brief introduction of document under the title.   \\\hline
    MD5 of Document & String & A string for identifying the document.   \\\hline
    Ranking Position & Discrete Number & The document's displaying order on the screen. \\\hline   
    Multimedia Type & Discrete Number & The type of url, for example, advertisement, videos, maps.   \\\hline
    SERP to Top & Discrete Number & The vertical pixels of the SERP to the top of the screen. \\\hline 
    SERP Height & Discrete Number & The vertical pixels of SERP on the screen.  \\
    \toprule
    \end{tabular}
}
    \label{tab:page_presentation}
\end{table}
}

\begin{table}[h]
    \caption{Rich User Behaviors in \dataset{}}
    \centering
    \tabcolsep 0.02in
    \scalebox{0.82}{
    \begin{tabular}{l|l|l}
    \bottomrule
    User Behaviors & Feature Type & Explanation      \\\hline
    Query Reformulation & \makecell[l]{Sequential desensitization \\  tokens.} &\makecell[l]{The subsequent queries issued by users under the same\\ search goal. A session can have multiple queries.} \\\hline
    Click & Discrete Number & whether user clicked the document. \\\hline
    First Click & Discrete Number & The identifier of users' first click in a query. \\\hline
    Skip & Discrete Number & whether user skipped the document on the screen. \\\hline
    Final Click & Discrete Number & The identifier of users' last click in a query session. \\\hline 
    Slipoff Count & Discrete Number & The count of document being sliped off the screen. \\\hline
    Slipoff Count After Click & Discrete Number & The count of slipoff after user click the document. \\\hline
    Dwelling Time & Continuous Value & \makecell[l]{The length of time a user spends looking at a document \\ after they've clicked a link on a SERP page, but before \\ clicking back to the SERP results.} \\\hline
    Displayed Time & Continuous Value & document display time on the screen. \\ \hline
    Displayed Count & Discrete Number & The document's display count on the screen. \\\hline
    Displayed Time on Top Screen & Continuous Value & The document's display time on the top 1/3 of screen. \\\hline
    Displayed Count on Top Screen & Discrete Number & The document's display count on the top 1/3 of screen. \\ \hline
    Displayed Time on Middle Screen & Continuous Value & The document's display time on the middle 1/3 of screen. \\\hline
    Displayed Count on Middle Screen & Discrete Number & The document's display count on the middle 1/3 of screen. \\\hline
    Displayed Time on Bottom Screen & Continuous Value & The document's display time on the bottom 1/3 of screen. \\\hline
    Displayed Count on Bottom Screen & Discrete Number & The document's display count on the bottom 1/3 of screen. \\\hline
    Last Click & Discrete Number & The identifier of users' last click in a query. \\\hline     
    Reverse Display Count & Discrete Number & \makecell[l]{The document's display count of user view with\\ a reverse browse order from bottom to the top.} \\\hline  
    \toprule
    \end{tabular}}
    \label{tab:user_behavior}
\end{table}

\section{Further data analysis}\label{app:data_analysis}
In this section, we provide detailed data analysis results corresponding to the data analysis conclusion in Sec. \ref{sec:data_analysis}.

We briefly mention the data analysis results again as follows:
\begin{enumerate}
    \item Long-tail distribution appears in many user behaviors. 
    \item The logged search results on the high-frequency queries are more relevant than those on tail queries.
    \item The most important user feedback, click, shows a strong correlation with both page presentation feature, position, and other user behaviors, displayed count. 
    \item As the displayed time becomes longer, users are more likely to spend more time on the top of the screen.
\end{enumerate}

\begin{figure*}
\begin{minipage}{.49\textwidth}
\vspace{0mm}
\hspace{-5mm}
\begin{center}
    \includegraphics[width=1.0 \textwidth]{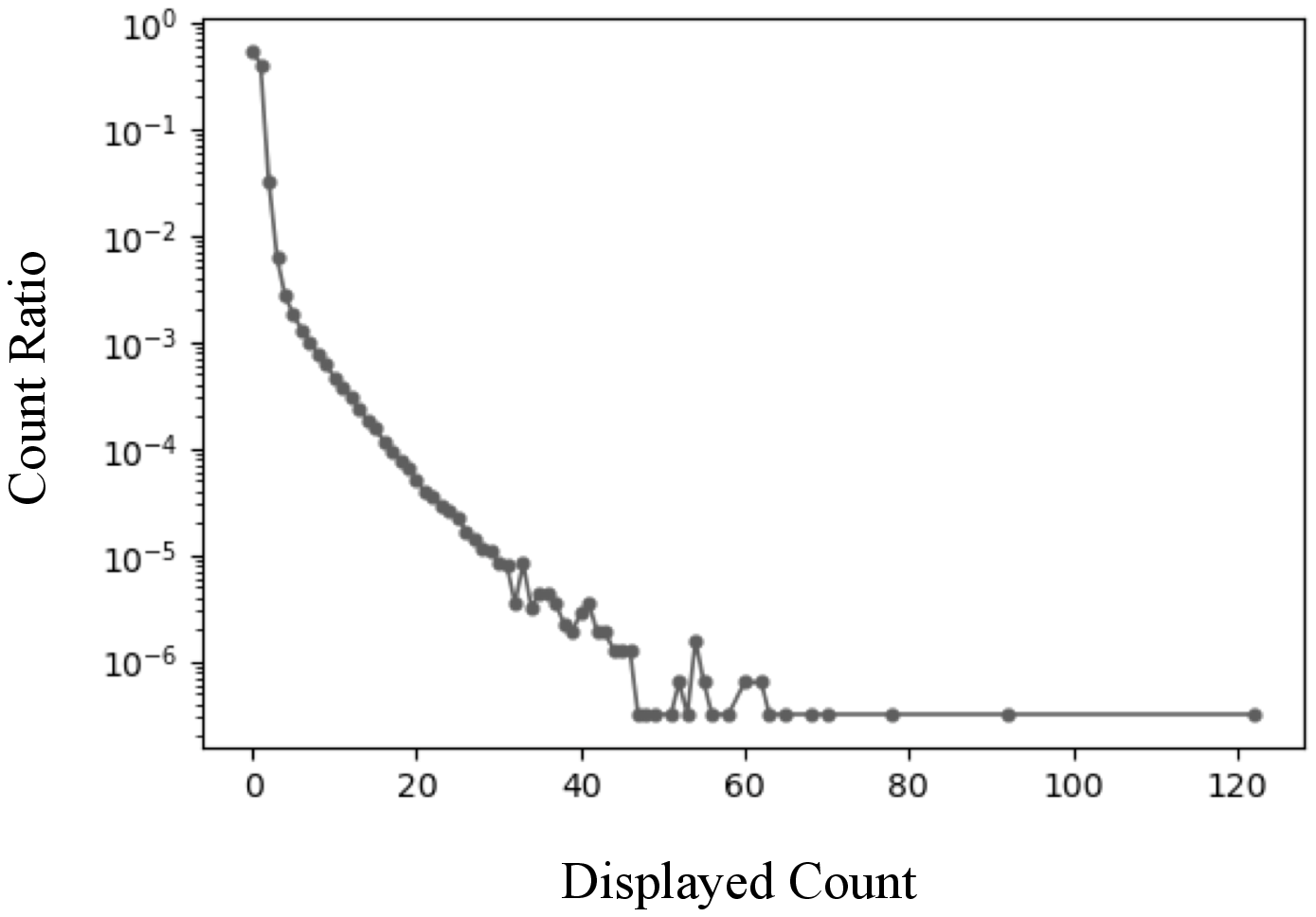}
    \end{center}
    \caption{The distribution of displayed counts. }
    \label{fig:app_view_count}
\end{minipage}
\begin{minipage}{.49\textwidth}
\vspace{-10mm}
\hspace{-10mm}
\begin{center}
    \includegraphics[width=0.95\textwidth]{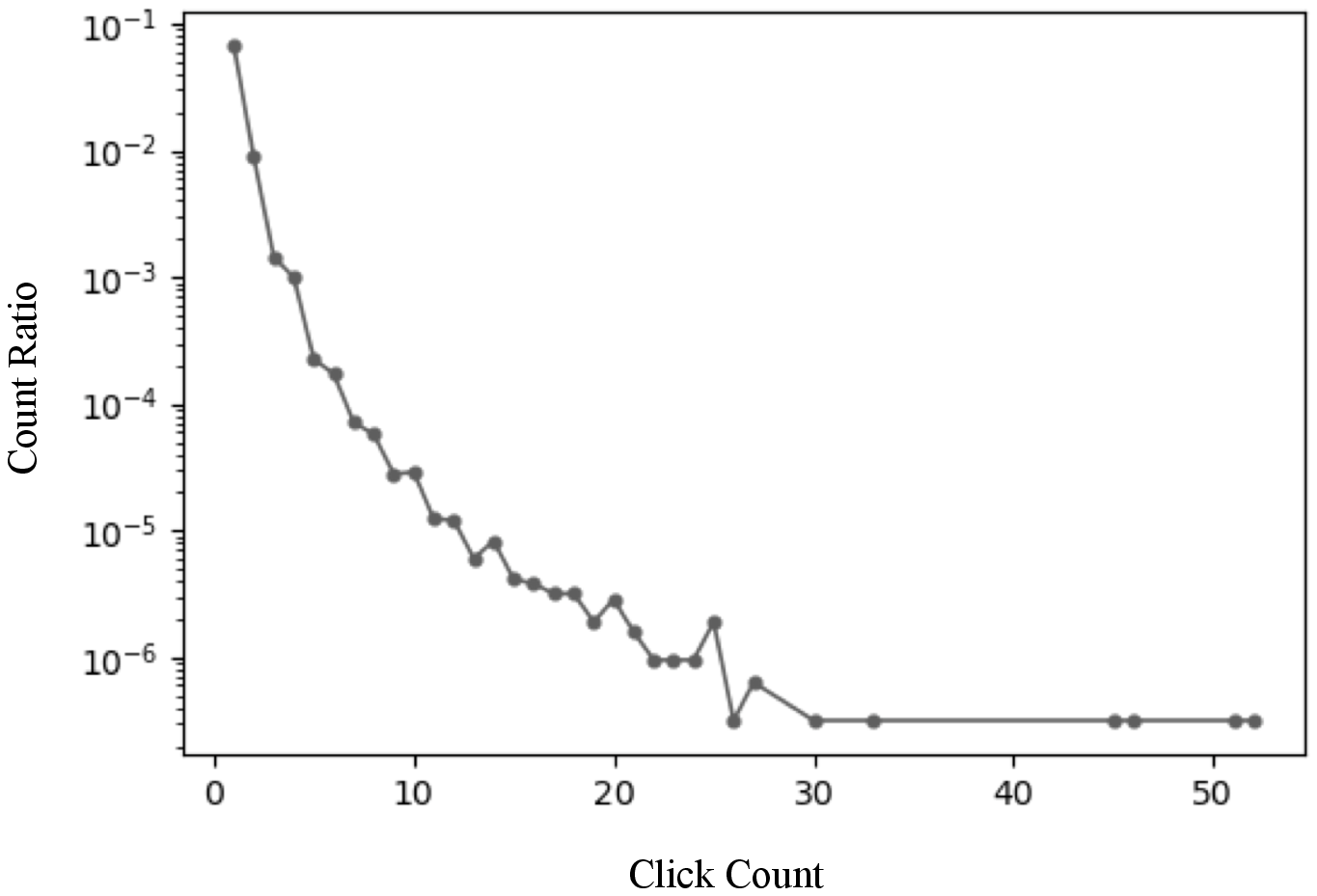}
    \end{center}
    \caption{The distribution of click count.}
    \label{fig:app_click_count}
    \vspace{-3mm}
\end{minipage}
\end{figure*}

\begin{figure*}
\begin{minipage}{.49\textwidth}
\vspace{0mm}
\hspace{-5mm}
\begin{center}
    \includegraphics[width=1.0 \textwidth]{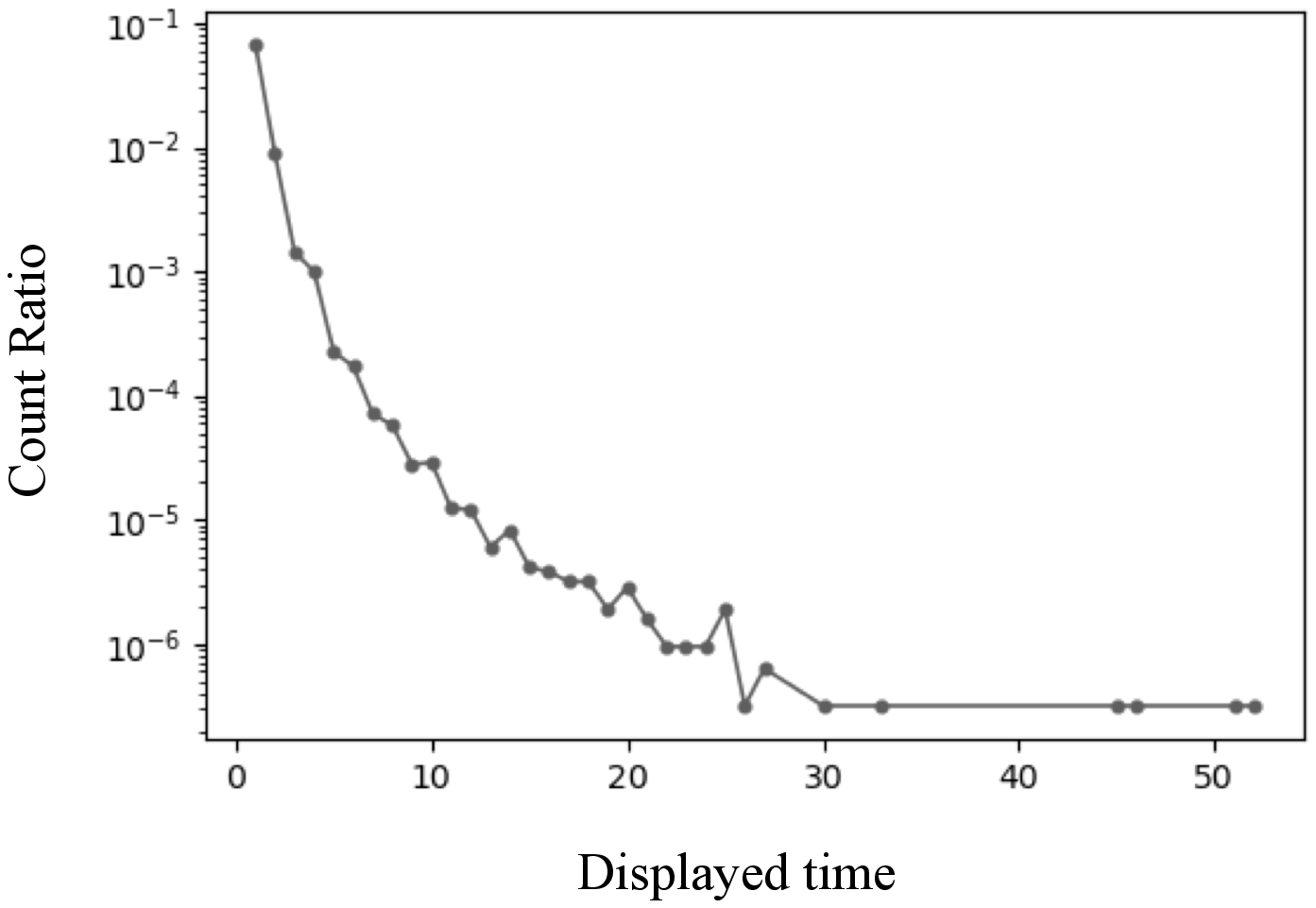}
    \end{center}
    \caption{The distribution of displayed time. }
    \label{fig:app_view_time}
\end{minipage}
\begin{minipage}{.49\textwidth}
\vspace{2mm}
\hspace{5mm}
\begin{center}
    \includegraphics[width=0.95\textwidth]{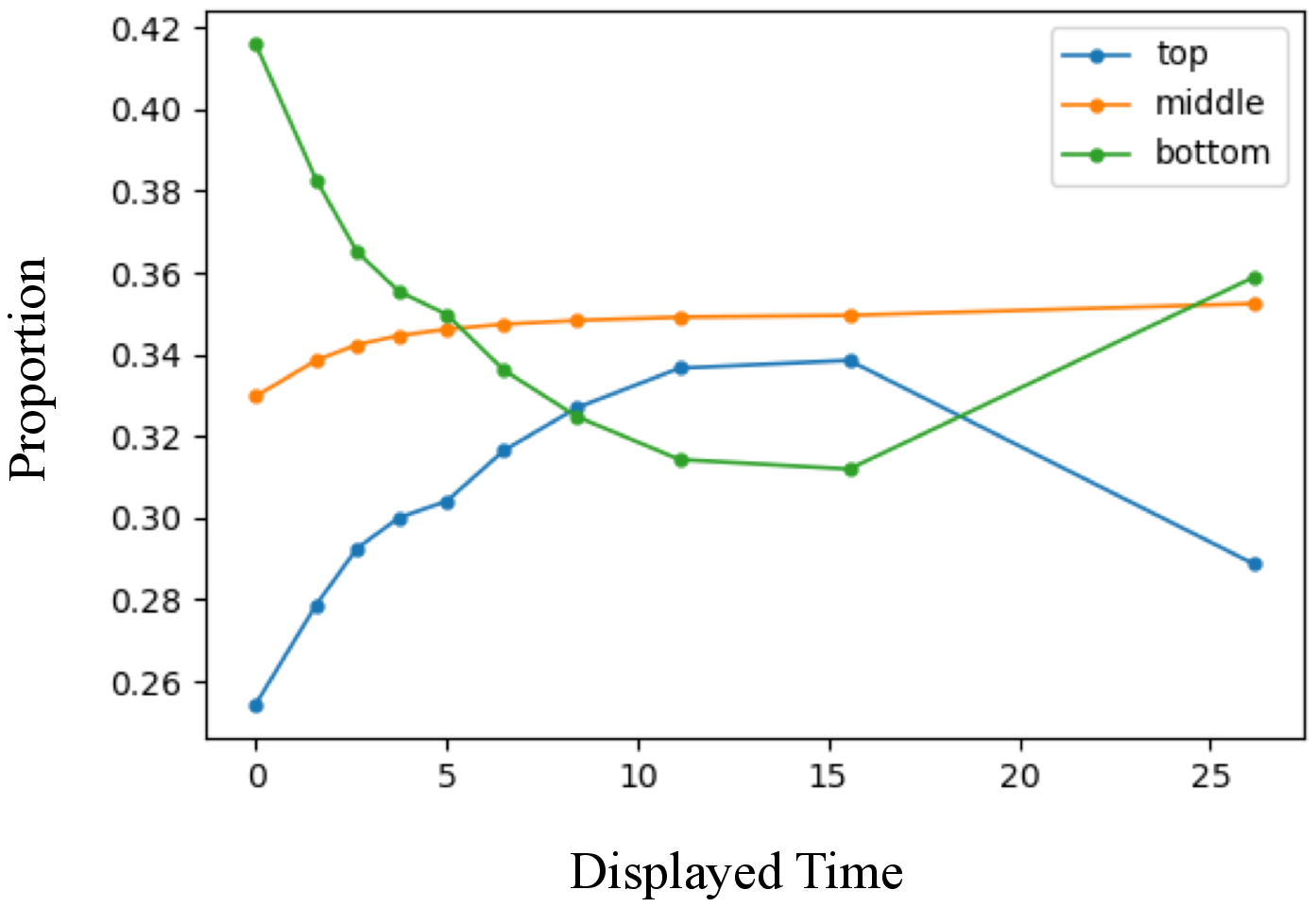}
    \end{center}
    \caption{Displayed time ratio on different parts of the screen.}
    \label{fig:app_different_screen}
    \vspace{-3mm}
\end{minipage}
\end{figure*}

Our supplementary focuses more on providing detailed results on conclusion 1 and conclusion 4.
For conclusion 2 and conclusion 3, all analysis results are included in Sec.~\ref{sec:data_analysis}.
For the conclusion 1 on long-tail distribution, we illustrate the distribution of displayed count, the displayed time, click count in Fig. \ref{fig:app_view_count}, Fig. ~\ref{fig:app_view_time}, Fig. ~\ref{fig:app_click_count}, respectively. It is obvious that all those data distributions reveal a clear long tail phenomenon.
Notice that for the continuous feature display time, we discretize the feature by splitting the interval 0.5 into the same bucket and calculate the count in each bucket.

For conclusion 4, the detailed data analysis results are shown in Fig. \ref{fig:app_different_screen}.
The analysis procedure is as follows.
We first remove documents with zero view time, which occupies a proportion of 54.7\%.
Then we ascendingly split data into 10 buckets according to the view time length, where each bucket has the same amount of documents.
For each bucket, we calculate the proportion of document view time on the top, middle, bottom of the screen. 
For example, the proportion on the top of the screen is calculated by top displayed time / (top displayed time + middle displayed time + bottom displayed time). 
We can see that, as the displayed time becomes longer, users are more likely to spend more time on the top of the screen, from 25\% to nearly 35\%,  while less time is spent on the bottom of the screen from more than 40\% to 30\%.

\end{document}